\documentclass[letterpaper, 10 pt, journal, twoside]{IEEEtran}
\usepackage{amsmath,amssymb,stmaryrd,mathtools}
\usepackage[hidelinks]{hyperref}
\usepackage[noabbrev]{cleveref}
\usepackage{cite}
\usepackage{bm}
\usepackage{booktabs}
\usepackage{multirow}
\usepackage{makecell}
\usepackage{xcolor}
\usepackage{paralist}
\usepackage{subcaption}
\usepackage{caption}
\begin{document}

\title{W-RIZZ: A Weakly-Supervised Framework for Relative Traversability Estimation in Mobile Robotics}

\author{Andre Schreiber, Arun N. Sivakumar, Peter Du, Mateus V. Gasparino, \\ Girish Chowdhary, and Katherine Driggs-Campbell%
\thanks{Manuscript received: January 23, 2024; Revised April 3, 2024; Accepted April 14, 2024.}
\thanks{This paper was recommended for publication by Editor Pascal Vasseur upon evaluation of the Associate Editor and Reviewers' comments.
This work was supported in part by the National Robotics Initiative 2.0 (NIFA\#2021-67021-33449), AIFARMS through the Agriculture and Food Research Initiative (AFRI) grant no. 2020-67021-32799/project accession no.1024178 from the USDA/NIFA, and UIUC's Center for Autonomous Construction in Manufacturing at Scale.
The robot platforms and data were provided by the Illinois Autonomous Farm and the Illinois Center for Digital Agriculture.} 
\thanks{Andre Schreiber, Arun N. Sivakumar, Peter Du, Mateus V. Gasparino, Girish Chowdhary, and Katherine Driggs-Campbell are with Coordinated Science Laboratory, University of Illinois at Urbana-Champaign, Champaign IL, 61820 USA.
        {\tt\footnotesize \{andrems2,av7,peterdu2,mvalve2,girishc,krdc\} \\@illinois.edu}}%
\thanks{Digital Object Identifier (DOI): 10.1109/LRA.2024.3396095.}
}

\markboth{IEEE Robotics and Automation Letters. Preprint Version. Accepted April, 2024}
{Schreiber \MakeLowercase{\textit{et al.}}: A Weakly-Supervised Framework for
Relative Traversability Estimation in Mobile Robotics} 

\maketitle

\begin{abstract}
Successful deployment of mobile robots in unstructured domains requires an understanding of the environment and terrain to avoid hazardous areas, getting stuck, and colliding with obstacles. Traversability estimation--which predicts where in the environment a robot can travel--is one prominent approach that tackles this problem. Existing geometric methods may ignore important semantic considerations, while semantic segmentation approaches involve a tedious labeling process. Recent self-supervised methods reduce labeling tedium, but require additional data or models and tend to struggle to explicitly label untraversable areas. To address these limitations, we introduce a weakly-supervised method for relative traversability estimation. Our method involves manually annotating the relative traversability of a small number of point pairs, which significantly reduces labeling effort compared to traditional segmentation-based methods and avoids the limitations of self-supervised methods. We further improve the performance of our method through a novel cross-image labeling strategy and loss function. We demonstrate the viability and performance of our method through deployment on a mobile robot in outdoor environments.
Code is available at: \url{https://github.com/andreschreiber/W-RIZZ}.
\end{abstract}


\IEEEpeerreviewmaketitle

\section{INTRODUCTION}
\IEEEPARstart{W}{ith} the rapidly expanding use of mobile robots in areas like agriculture \cite{ag_embedded} and delivery \cite{deliver_robot}, it is critical that such robots are able to effectively understand their environments in order to navigate and successfully complete their tasks. 
Traversability estimation, which involves determining where in the environment a robot can travel, is one common method for solving this problem in unstructured environments \cite{papadakis2013terrain}. Due to advances in computer vision and visual perception systems in recent years, vision-based methods for traversability estimation have shown significant promise~\cite{wayfast, wherewalk, selfsuponly, reconsdrive, catcavs}.

Vision-based methods for environmental understanding and traversability estimation can typically be classified into strongly-supervised methods based on semantic segmentation and self-supervised methods. Strongly-supervised semantic segmentation approaches~\cite{catcavs, adapnet, freiburg, maturana, rugd, rellis3d} involve using a semantic segmentation model to segment the environment into classes relevant to navigation and traversability. However, these methods suffer from a tedious and expensive manual labeling process as training images are densely annotated by a human labeler. Self-supervised approaches avoid this labeling burden by annotating data automatically \cite{wayfast, wherewalk, selfsuponly, reconsdrive}, but typically involve additional assumptions, data, or models. In addition, these self-supervised approaches generally require that the robot experience conditions of interest to label them, which creates difficulty in explicitly labeling untraversable conditions in a safe manner \cite{selfsuponly, reconsdrive}.

\begin{figure}
  \centering
  \includegraphics[width=3.35in]
  {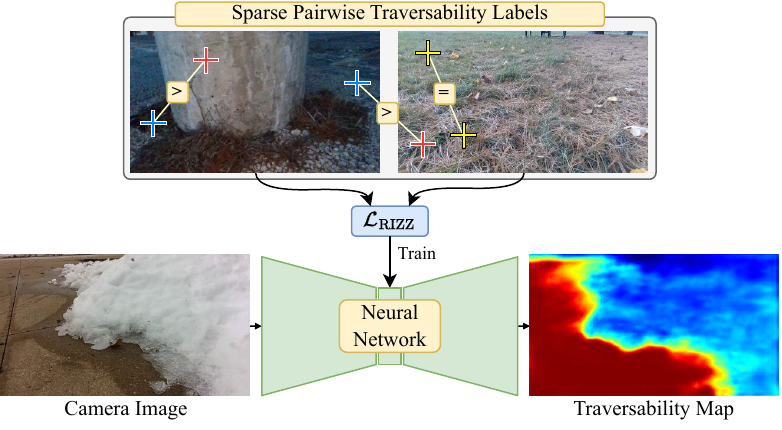}
  \vspace{-2pt}
  \caption{Our proposed framework trains a model for traversability estimation using sparse pairwise labels of relative traversability. The camera images are from the dataset introduced by Gasparino \textit{et al.} \cite{wayfast}. In the traversability map, warmer colors indicate higher anticipated ease of traversal.}
  \label{fig_visual_abstract}
  \vspace{-15pt}
\end{figure}

Motivated by the limitations of these existing approaches, we propose a weakly-supervised framework for visual traversability estimation, termed Weak-Relative Inference of haZard Zones (W-RIZZ). We draw inspiration from work in relative depth estimation \cite{depthwild,zoran,snow} and intrinsic decomposition \cite{intrinsic,zoran}, and propose annotating sparse pairs of points with relative traversability labels (as shown in Fig. \ref{fig_visual_abstract}). The sparse labeling greatly reduces the tedium seen in strongly-supervised methods, while enabling easy explicit labeling of untraversable regions and not requiring robot experience during training. Our method can also provide continuous predictions of traversability (rather than discrete classes \cite{catcavs}), and does not require the additional assumptions, data, or models needed in self-supervised methods \cite{wayfast, selfsuponly, wherewalk, reconsdrive}.

We summarize our contributions as follows:
\begin{enumerate}
    \item We introduce a weakly-supervised framework for training traversability estimation neural networks that employs sparse pairwise labels of relative traversability.
    \item We propose a new loss for training traversability estimation models and a novel cross-image pairwise labeling scheme that improves consistency of traversability predictions across different images and environments.
    \item We deploy our approach on a robot and demonstrate that our model outperforms a state-of-the-art self-supervised method for visual traversability prediction, showing greater generalization capability and improved navigation success rate while only using a small number of annotated pixels per image.
\end{enumerate}

\section{Related Works}

Numerous potential approaches to mobile robot navigation and traversability estimation have been proposed and are relevant to the work presented in this letter.

\subsection{Geometric and Semantic Approaches to Navigation}
A classical approach to mobile robot navigation involves the creation of an environmental map using sensors like LiDAR, localizing the robot in the map using SLAM, and planning an obstacle-free path through the map \cite{probabilisticrobotics,siegwart}. However, only considering geometry in navigation can be problematic since it fails to consider non-rigid and traversable obstacles like tall grass and may ignore characteristics like the bumpiness of a traversed surface.

These limitations can be partially resolved by integrating semantic information about the environment. For example, Valada \textit{et al.} \cite{adapnet} introduce a robust segmentation architecture that uses a convoluted mixture of deep experts model. Maturana \textit{et al.} \cite{maturana} utilize geometric and semantic information, where environment semantics are predicted using a neural network and semantic classes are assigned navigation costs.

Several datasets exist for semantic segmentation in outdoor unstructured environments. For example, the Robot Unstructured Ground Driving (RUGD) dataset \cite{rugd} and Freiburg Forest dataset \cite{freiburg} provide images that are densely labeled with semantic classes relevant for outdoor navigation. The CaT: CAVS \cite{catcavs} dataset features images collected from an off-road car, and directly labels regions as traversable by a specific vehicle platform. However, these datasets only contain a few hundred \cite{freiburg} to a few thousand \cite{catcavs,rugd} images, which is significantly smaller than traditional datasets for segmentation in computer vision \cite{cocodataset}.

Methods based on semantic segmentation can be appealing as they can provide more fine-grained understanding than purely geometric approaches. Moreover, numerous architectures for semantic segmentation exist (e.g., PSPNet \cite{pspnet}) and can be implemented for the task of semantic segmentation in unstructured environments.
However, densely labeling datasets for segmentation is tedious, making it difficult to apply this method in a new domain. In addition, choosing an appropriate set of semantic classes may be challenging and can vary for different applications. It may also be difficult to map these semantic classes into costs that can be used for control and motion planning~\cite{howdoesitfeel}.
Segmenting images directly for traversability can also be problematic, as traversability segmentation labels typically represent only binary indications of traversability and may only apply to a specific robotic platform~\cite{catcavs}.
Finally, each of these forms of semantic segmentation fails to consider the possibility of traversability characteristics varying within semantic classes.

\subsection{Self-Supervised Navigation and Traversability}
To overcome the limitations of geometric and segmentation approaches to mobile robot navigation, much recent work has focused on developing self-supervised methods for navigation and traversability estimation. BADGR \cite{badgr} collects and labels off-policy data using a simple navigation policy, and trains an action-conditioned neural network to predict navigation events. However, BADGR requires a large dataset to perform well, as it learns both the navigation characteristics of the environment and dynamics of the robot. 

Other methods \cite{wayfast,probtrav} overcome this limitation by directly incorporating a dynamics model. For example, WayFAST \cite{wayfast} uses the kinodynamic model of a skid-steer robot to measure traction coefficients. These traction coefficients are then projected into camera images and used to train a traversability prediction neural network. For quadrupeds, Wellhausen \textit{et al.}~\cite{wherewalk} propose projecting footholds into images and retroactively assign labels using proprioceptive measurements.
Similarly, Castro \textit{et al.} \cite{howdoesitfeel} combine proprioceptive sensing of terrain interactions with exteroceptive sensor data to learn costmaps for off-road driving.
Other proposed methods involve labeling traversed paths as traversable and training a classifier on such paths \cite{selfsuponly}, or training an autoencoder only on traversed regions and using reconstruction error as a proxy for traversability \cite{reconsdrive}.

These self-supervised methods are appealing by eliminating the burden of manual labeling through self-supervision. However, such methods are typically only weakly-supervised (they do not label all training image pixels) and can be difficult to apply on existing image-only datasets as they often require additional data sources. The need for the robot to experience conditions to label them also creates difficulty in labeling non-traversable areas, since subjecting the robot to such conditions may not be possible or can lead to damage. As a result, these methods may assume unfamiliar regions as non-traversable \cite{reconsdrive} and are often biased towards a positive and unlabeled (PU) \cite{selfsuponly,scate} data regime.

\subsection{Weakly-Supervised Learning}
Learning with sparse labels (weakly-supervised learning) is also heavily related to our proposed method. Various works have proposed methods for reducing the labeling burden in semantic segmentation with approaches using image-level labels~\cite{imagelevel} or using annotations such as points~\cite{point}  or squiggles~\cite{squiggle}.
For mobile robot navigation in particular, Gao \textit{et al.}~\cite{contrastivetrav} introduce a contrastive learning approach for traversability estimation, and Schreiber~\cite{schreiber2023weakly} presents a method for weakly-supervised traversability segmentation using point annotation. Methods have also been introduced for boosting the accuracy of weakly-supervised models~\cite{temporalensembling,meanteacher} and have successfully been applied in field robotics~\cite{wherewalk, schreiber2023weakly}.

Beyond semantic segmentation, weakly-supervised learning has been utilized with pairwise labels for tasks like intrinsic image decomposition~\cite{intrinsic,zoran} and depth estimation~\cite{zoran,depthwild,snow}. In these methods, relative annotations for pairs of points are provided and are used with ranking-based losses to produce continuous outputs (in contrast to the discrete class outputs seen in segmentation). We draw inspiration from such pairwise labeling schemes and adapt this idea to weakly-supervised traversability prediction.

\section{Method}

We propose a framework (W-RIZZ) for weakly-supervised relative traversability estimation. The proposed framework involves manually labeling a small number of point pairs in each training image, where one point in a pair is labeled as more, less, or equally traversable as compared with the other point in the pair. By using sparse annotations, the labeling burden compared to a strongly-supervised approach is greatly reduced, with the time to annotate an image in W-RIZZ taking seconds rather than minutes. The use of sparse manual annotations (rather than self-supervision) means that our method does not require any additional data, models, or assumptions and can be used on existing image-only datasets. Finally, the lack of self-supervision in W-RIZZ means that untraversable regions can be labeled explicitly without risking damage to the robot or its environment.

\subsection{Data Annotation}
Inspired by relative annotation strategies for tasks like depth estimation \cite{depthwild,zoran,snow} and intrinsic decomposition \cite{intrinsic,zoran}, we label images using a small number of pairwise annotations of sparse points. Specifically, given an image to be annotated, a labeler is prompted with a pair of points and asked to indicate which point in the pair is more traversable (or if the points are equally traversable). The locations of the points in a pair are selected randomly in order to capture the natural traversability statistics of images in the dataset and since  random sampling has been shown to be particularly effective in weakly-supervised semantic segmentation \cite{point}.

If relative traversability labels are only provided for pairs where both pixels belong to the same image, the scale of the traversability predictions across different images may vary. This can lead to inconsistent predictions and may require tweaks to the controller that uses the traversability predictions. Thus, unlike the relative pairwise labeling strategies described in other works (which only label relationships within the same image) \cite{depthwild,zoran,intrinsic,snow}, we propose to also label pairs where each point belongs to a different image.

\begin{figure}[tp]
  \centering
  \includegraphics[width=3in]{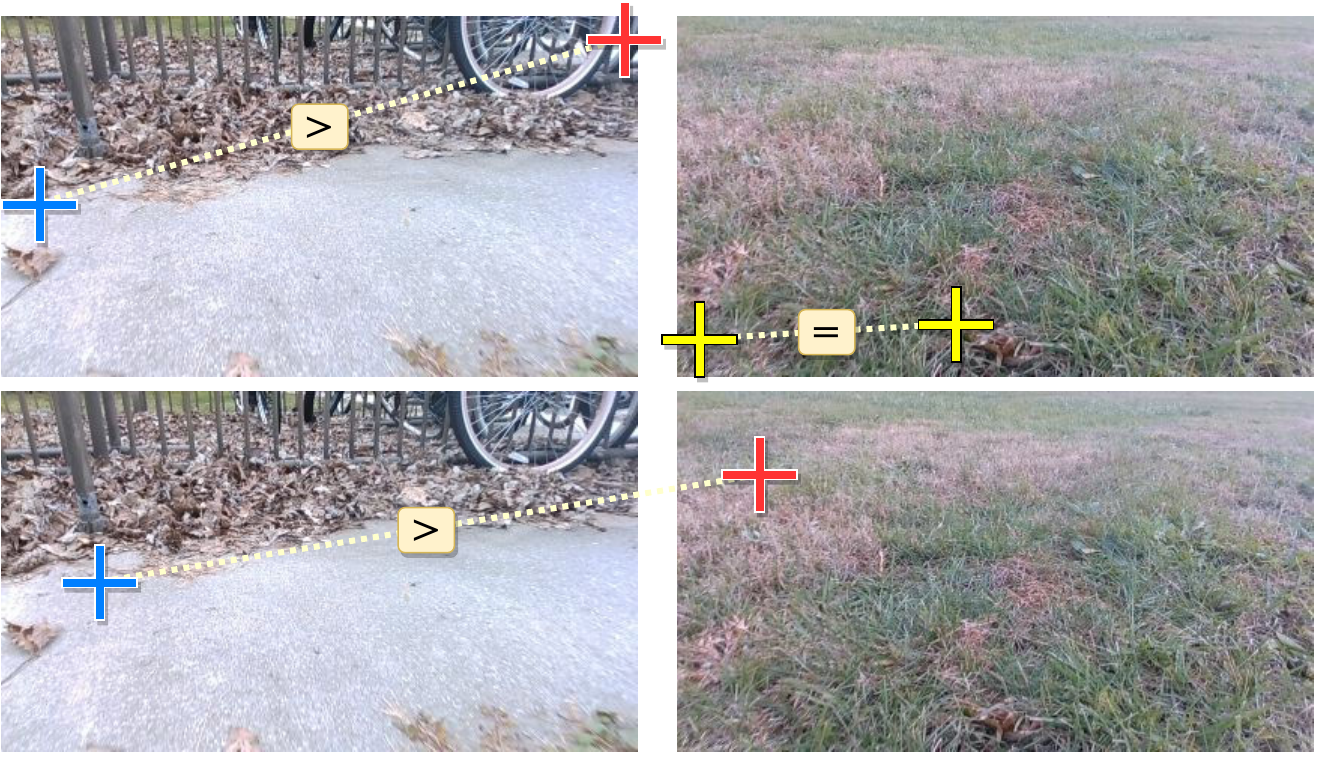}
  \caption{Illustration of our annotation strategy on the WayFAST dataset \cite{wayfast}, showing intra-image labeling (top) and cross-image labeling (bottom). Yellow crosshairs indicate both points in a pair are equally traversable, while the blue point is more traversable in pairs with blue and red crosshairs.}
  \label{fig_labeling}
  \vspace{-15pt}
\end{figure}

For each image in a dataset, we label the relative traversability of one pair of pixels within the image (an ``intra-image'' label), and we also provide a ``cross-image'' label (where one pixel belongs to the image and the other pixel in the pair belongs to a different image). The pairings for cross-image labels are chosen randomly, and one cross-image label is provided per image. The annotation strategy leads to 3 pairwise labels for every 2 images (1 intra-image pair label for each image and 1 cross-image pair label). For intra-image labels, we limit selected pixels for an annotation pair to be at least 5\% of $\text{min}(W,H)$ apart (where $W$ and $H$ are the width and height of the image) to avoid labeling pairs where the pixels represent nearly the same location. Fig.~\ref{fig_labeling} illustrates this labeling strategy on example images from the dataset introduced by Gasparino \textit{et al.} \cite{wayfast}. Labeling the 16558 images from this dataset with our annotation method took approximately 3 standard working days and produced 24837 pairwise annotations.

\subsection{Loss Function}
We train a model in our proposed framework using ordinal annotations of relative traversability. While various works describe methods for learning vision tasks using relative relationships \cite{intrinsic, zoran, depthwild,snow}, the method of Chen \textit{et al.} \cite{depthwild} is most similar to our desired application, presenting a ranking loss specifically for end-to-end training of depth estimation neural networks using sparse pairwise annotations.
The loss introduced by Chen \textit{et al.} \cite{depthwild} can be described as follows:
\begin{equation} \label{eq1}
\mathcal{L}_{\text{DIW}}(p_a,p_b,t)=\begin{cases} 
    \text{ln}(1+\text{exp}(-t(p_b-p_a))) & t \not = 0 \\
    (p_b-p_a)^2 & t = 0 \\
\end{cases}
\end{equation}
where $p_a$ is the prediction value of the first point in the pair (point $a$), $p_b$ is the prediction value of the second point in the pair (point $b$), and $t \in \{-1,0,1\}$ is the ordinal annotation of the point pair (with $t=0$ being an equality relation, $t=1$ meaning $p_b$ should be greater than $p_a$, and $t=-1$ meaning $p_b$ should be less than $p_a$).

In the case of inequality labels ($t\not=0$), the $\mathcal{L}_\text{DIW}$ loss features a term reminiscent of the log-loss, which pushes prediction values in inequality pairs to be infinitely far apart. For our task of relative traversability prediction, we found that this leads to traversability predictions mostly at the edges of the prediction range, which is unattractive as it degrades the predictions to binary (go/no-go) traversability prediction.

Later work in relative depth and surface normal estimation also describes this problem of the loss encouraging predictions in inequality pairs to be infinitely far apart \cite{snow}, and proposes a modified loss ($\mathcal{L}_\text{SNOW}$) that clamps the difference $p_b-p_a$ in the original $\mathcal{L}_\text{DIW}$ loss to remedy the issue. We instead propose to introduce a new loss ($\mathcal{L}_{\text{RIZZ}}$) based on a squared hinge loss. Our use of the squared hinge loss allows for squared penalty for both inequality and equality relations, while also resolving the issue of producing mostly extreme predictions as the hinge loss is zero once predictions are sufficiently correct. We have found that this loss provides improved performance over both $\mathcal{L}_\text{DIW}$ and $\mathcal{L}_\text{SNOW}$.
Our $\mathcal{L}_{\text{RIZZ}}$ loss is calculated as follows:
\begin{equation} \label{eq2}
\mathcal{L}_{\text{RIZZ}}(p_a,p_b,t)=\begin{cases} 
    \text{max}\{0,L-t(p_b-p_a)\}^2 & t \not = 0 \\
    (p_b-p_a)^2 & t = 0 \\
\end{cases}
\end{equation}
where $p_a$, $p_b$, and $t$ are defined as before, and $L$ is a margin hyperparameter.

\subsection{Training and Network Architecture}
\label{sec:method_arch}
We train a model in our framework using our proposed $\mathcal{L}_{\text{RIZZ}}$ loss. However, like other work in weakly-supervised learning for field robotics \cite{wherewalk,schreiber2023weakly}, we use mean teacher \cite{meanteacher} to improve the performance of our model when trained with sparse annotations. We maintain a teacher network and a student network that share the same architecture, where the teacher network weights are an exponential moving average of the student network weights.
The final loss used for training a model with W-RIZZ is a weighted sum of an accuracy-based loss ($\mathcal{L}_{\text{RIZZ}}$) and a mean teacher consistency loss (mean squared error). The accuracy-based loss is computed with the student network predictions and is used to learn the relative traversability score, while the consistency loss penalizes differences between the teacher and student network predictions.

Our method is agnostic to the network architecture that is used. Any architecture for per-pixel regression tasks can be used within the W-RIZZ framework, and different models can be selected to balance accuracy and inference speed.
For the experiments in this letter, we adopt a modified version of the TravNet architecture introduced by Gasparino \textit{et al.}~\cite{wayfast}. We only use RGB camera images as input and we keep the same encoder (based on ResNet-18 \cite{resnet}). We modify the TravNet decoder by replacing the transpose convolutional layers with standard $3 \times 3$ convolution layers that are followed by nearest-neighbor interpolation, and as our final layer we use a $1 \times 1$ convolutional layer with a sigmoid activation to yield a traversability score between 0 and 1 for each pixel. The network can be viewed as a function $f_\theta: \mathbb{R}^{H \times W \times 3} \mapsto [0,1]^{H \times W}$. We modified the decoder to reduce checkerboarding artifacts seen in the predictions of the original TravNet.

In training, we apply data augmentations in the form of color jitter, random horizontal flipping, and random cropping. As described by Wellhausen \textit{et al.} \cite{wherewalk}, geometric augmentations (cropping and flipping) are applied in the same way to the inputs for the teacher and student network, while color jitter is sampled independently for the teacher and student.

\section{Experimental Results}
We validate our method using the WayFAST dataset \cite{wayfast}. The dataset contains 16558 images across a variety of conditions, such as snow, tall grass, forest-like terrain, and semi-urban areas. We label the dataset with 24837 annotations of relative traversability (3 pairwise annotations for every 2 images) for a total of 49674 labeled pixels for the dataset.

For our ablation studies, we split the dataset into a training and validation split, with 13248 training images and 3310 validation images. In all of our experiments, we use a resolution of $H \times W = 240 \times 424$ for the inputs and outputs of our network. Our dataset contains more equality ordinal labels than inequality labels, and we oversample inequality labels to achieve a more balanced number of labels.  For all offline results, we use a computer with an RTX 3080 GPU and neural network inference takes $3 \ \text{ms}$ per image.

\subsection{Quantitative Results}\label{sec:quant_results}

We measure the quantitative performance of our model using Human Disagreement Rate (HDR), which is an adaptation of the Weighted Human Disagreement Rate (WHDR) metric given by Zoran \textit{et al}. \cite{zoran}:
\begin{equation} \label{eq6}
\text{HDR}_{\tau}(p,t) = \frac{\sum_{i}^{N} \mathbf{1}(L_{\tau}(p_{i,a}, p_{i,b}) \not = t_i) }{N}
\end{equation}
where $p \in [0,1]^{N \times 2}$ is the set of neural network predictions at the labeled points, $t \in \{-1,0,1\}^N$ is the set of ground-truth ordinal labels, and $N$ is the total number of labels. $L_\tau$ maps prediction pairs to ordinal labels according to threshold $\tau$ and is defined as follows:
\begin{equation} \label{eq7}
L_{\tau}(p_a, p_b)=\begin{cases}
    1, & p_b - p_a > \tau \\
    0, & |p_b - p_a| \leq \tau \\
    -1, & p_b - p_a < -\tau \\
\end{cases}
\end{equation}

We analyze the following model variants: a model with no pretraining on RUGD~\cite{rugd} and no mean teacher~\cite{meanteacher} (\textit{INIT}); a model using mean teacher but with no pretraining on RUGD (\textit{MT}); and a model using mean teacher and that is pretrained on RUGD (\textit{MT+PT}).
For pretraining, we train our network for multi-class segmentation on RUGD and use the weights of all but the last layer to initialize our W-RIZZ model. We test the \textit{MT+PT} model with our $\mathcal{L}_{\text{RIZZ}}$ loss, as well as  $\mathcal{L}_{\text{DIW}}$ \cite{depthwild},  $\mathcal{L}_{\text{SNOW}}$ \cite{snow}, and $\mathcal{L}_{\text{RIZZ-L1}}$ (a variant of our loss that does not square the hinge loss and that uses absolute error for equality pairs). Any hyperparameters for losses are chosen via grid search (all experiments use a value of $L=0.5$ for $\mathcal{L}_{\text{RIZZ}}$).
In addition to the W-RIZZ variants, we analyze the performance of the self-supervised WayFAST~\cite{wayfast}, which is trained on the publicly available labels in the WayFAST dataset with the same network architecture used by W-RIZZ (which uses only RGB camera data and is pretrained on RUGD).

We show quantitative results on the validation set for $\text{HDR}_{\tau}$ (overall human disagreement rate), $\text{HDR}_{\tau}^{=}$ (human disagreement rate on pairs labeled as equality), and $\text{HDR}_{\tau}^{\not=}$ (human disagreement rate on pairs labeled as inequality) for our selected variants at a threshold of $\tau=0.25$ in Table I. In Table II, we provide results for WayFAST and the \textit{MT+PT} variant of W-RIZZ (with several different losses) at different values of the disagreement threshold $\tau$.

\begin{table}[bp]
\vspace{-15pt}
\caption{Results for Different Variants of W-RIZZ}
\vspace{-8pt}
\begin{center}
\begin{tabular}{@{} l *{4}{r} @{}}
\toprule
Model & \multicolumn{1}{c}{Loss} & \multicolumn{1}{c}{$\text{HDR}_{0.25}$} & \multicolumn{1}{c}{$\text{HDR}_{0.25}^{=}$} & \multicolumn{1}{c}{$\text{HDR}_{0.25}^{\not=}$}\\
\midrule
{INIT} & \multicolumn{1}{c}{$\mathcal{L}_{\text{RIZZ}}$} & \multicolumn{1}{c}{0.141} & \multicolumn{1}{c}{0.074} & \multicolumn{1}{c}{\textbf{0.266}} \\
{MT} & \multicolumn{1}{c}{$\mathcal{L}_{\text{RIZZ}}$} & \multicolumn{1}{c}{0.129} & \multicolumn{1}{c}{0.047} & \multicolumn{1}{c}{0.281} \\
{MT+PT} & \multicolumn{1}{c}{$\mathcal{L}_{\text{RIZZ}}$} & \multicolumn{1}{c}{\textbf{0.125}} & \multicolumn{1}{c}{\textbf{0.044}} & \multicolumn{1}{c}{0.275} \\
{MT+PT} & \multicolumn{1}{c}{$\mathcal{L}_{\text{DIW}}$} & \multicolumn{1}{c}{0.153} & \multicolumn{1}{c}{0.067} & \multicolumn{1}{c}{0.311} \\
{MT+PT} & \multicolumn{1}{c}{$\mathcal{L}_{\text{RIZZ-L1}}$} & \multicolumn{1}{c}{0.176} & \multicolumn{1}{c}{0.054} & \multicolumn{1}{c}{0.401} \\
{MT+PT} & \multicolumn{1}{c}{$\mathcal{L}_{\text{SNOW}}$} & \multicolumn{1}{c}{0.141} & \multicolumn{1}{c}{0.044} & \multicolumn{1}{c}{0.320} \\
\midrule
{WayFAST} & \multicolumn{1}{c}{} & \multicolumn{1}{c}{0.311} & \multicolumn{1}{c}{0.124} & \multicolumn{1}{c}{0.658} \\
\bottomrule
\end{tabular}
\end{center}
\vspace{-5pt}
\end{table}

In Table I, we see that the \textit{MT+PT} model with $\mathcal{L}_{\text{RIZZ}}$ performs best on $\text{HDR}_{0.25}$ and $\text{HDR}_{0.25}^{=}$ (and performs second best on $\text{HDR}_{0.25}^{\not=}$, showing only slightly higher error than the \textit{INIT} model). Such a result demonstrates that using our $\mathcal{L}_{\text{RIZZ}}$ loss along with mean teacher \cite{meanteacher} and pretraining on RUGD \cite{rugd} yields the best results. However, the \textit{MT} model (which is not pretrained on RUGD) performs only slightly worse than the \textit{MT+PT} model, indicating that pretraining using existing segmentation datasets is beneficial if possible but is not required for our approach to be successful.

The results in Table I and Table II show that our $\mathcal{L}_{\text{RIZZ}}$ loss outperforms the other losses across nearly all metrics. Only for $\text{HDR}_{0.5}$ (which involves a high threshold for equality) does $\mathcal{L}_{\text{RIZZ}}$ not perform best of all losses, and only $\mathcal{L}_{\text{DIW}}$ performs better on this metric. Moreover, $\mathcal{L}_{\text{DIW}}$ shows relatively similar error across equality thresholds, which suggests that it mostly produces predictions at the extremes of the prediction range (i.e., highly traversable or untraversable). In Table I and Table II, our method also shows significantly better $\text{HDR}$ than the self-supervised WayFAST \cite{wayfast} (which is trained with over $3000\times$ more labeled pixels), likely due to the added label noise introduced via self-supervised labeling. However, we do acknowledge that WayFAST is not specifically trained for optimizing human disagreement, so lower performance is somewhat expected.

We analyze the effect of the cross-image labeling in Table III. For this experiment, we train models using mean teacher, pretraining on RUGD, and $\mathcal{L}_{\text{RIZZ}}$ loss. We provide results when training using only intra-image labels (\textit{Intra}) and when training using cross-image and intra-image labels (\textit{Intra+Cross}). To ensure fair comparison, we train the two variants using the same number of annotations (so each method uses 2 labels for every 2 images). The results in Table III demonstrate that using cross-image and intra-image labels outperforms using solely intra-image labels both on overall $\text{HDR}_{0.25}$ and on $\text{HDR}_{0.25}^{\not=}$. On the other hand, we see that using only intra-image labels leads to slightly improved error on $\text{HDR}_{0.25}^{=}$ (although both labeling schemes show relatively low error on this metric). Nonetheless, the better overall $\text{HDR}_{0.25}$ indicates that using cross-image and intra-image labels is beneficial by significantly boosting the accuracy on pairs where the ground-truth ordinal relationship is inequality. Moreover, the use of cross-image and intra-image labels is beneficial as it produces more consistently calibrated traversability predictions (since predictions are optimized across images, not just within a given image).

\begin{table}[tp]
\vspace{5pt}
\caption{Results for Multiple Disagreement Thresholds}
\vspace{-8pt}
\begin{center}
\begin{tabular}{@{} l *{4}{r} @{}}
\toprule
{Method} & \multicolumn{1}{c}{Loss} & \multicolumn{1}{c}{$\text{HDR}_{0.1}$} & \multicolumn{1}{c}{$\text{HDR}_{0.25}$} & \multicolumn{1}{c}{$\text{HDR}_{0.5}$}\\
\midrule
{W-RIZZ} & \multicolumn{1}{c}{$\mathcal{L}_{\text{DIW}}$} & \multicolumn{1}{c}{0.193} & \multicolumn{1}{c}{0.153} & \multicolumn{1}{c}{\textbf{0.164}} \\
{W-RIZZ} & \multicolumn{1}{c}{$\mathcal{L}_{\text{SNOW}}$} & \multicolumn{1}{c}{0.195} & \multicolumn{1}{c}{0.141} & \multicolumn{1}{c}{0.266} \\
{W-RIZZ} & \multicolumn{1}{c}{$\mathcal{L}_{\text{RIZZ-L1}}$} & \multicolumn{1}{c}{0.182} & \multicolumn{1}{c}{0.176} & \multicolumn{1}{c}{0.293} \\
{W-RIZZ} & \multicolumn{1}{c}{$\mathcal{L}_{\text{RIZZ}}$} & \multicolumn{1}{c}{\textbf{0.165}} & \multicolumn{1}{c}{\textbf{0.125}} & \multicolumn{1}{c}{0.225} \\
\midrule
{WayFAST} & \multicolumn{1}{c}{} & \multicolumn{1}{c}{0.370} & \multicolumn{1}{c}{0.311} & \multicolumn{1}{c}{0.310} \\
\bottomrule
\end{tabular}
\end{center}
\vspace{-5pt}
\end{table}

\begin{table}[tp]
\caption{Intra- vs. Cross-Image Labeling}
\vspace{-8pt}
\begin{center}
\begin{tabular}{@{} l *{3}{r} @{}}
\toprule
Labeling Method & \multicolumn{1}{c}{$\text{HDR}_{0.25}$} & \multicolumn{1}{c}{$\text{HDR}_{0.25}^{=}$} & \multicolumn{1}{c}{$\text{HDR}_{0.25}^{\not=}$}\\
\midrule
Intra & \multicolumn{1}{c}{0.165} & \multicolumn{1}{c}{\textbf{0.025}} & \multicolumn{1}{c}{0.425} \\
Intra+Cross & \multicolumn{1}{c}{\textbf{0.133}} & \multicolumn{1}{c}{0.042} & \multicolumn{1}{c}{\textbf{0.301}} \\
\bottomrule
\end{tabular}
\end{center}
\vspace{-10pt}
\end{table} 

\begin{figure}[tp!]
  \centering
  \includegraphics[width=3.1in]{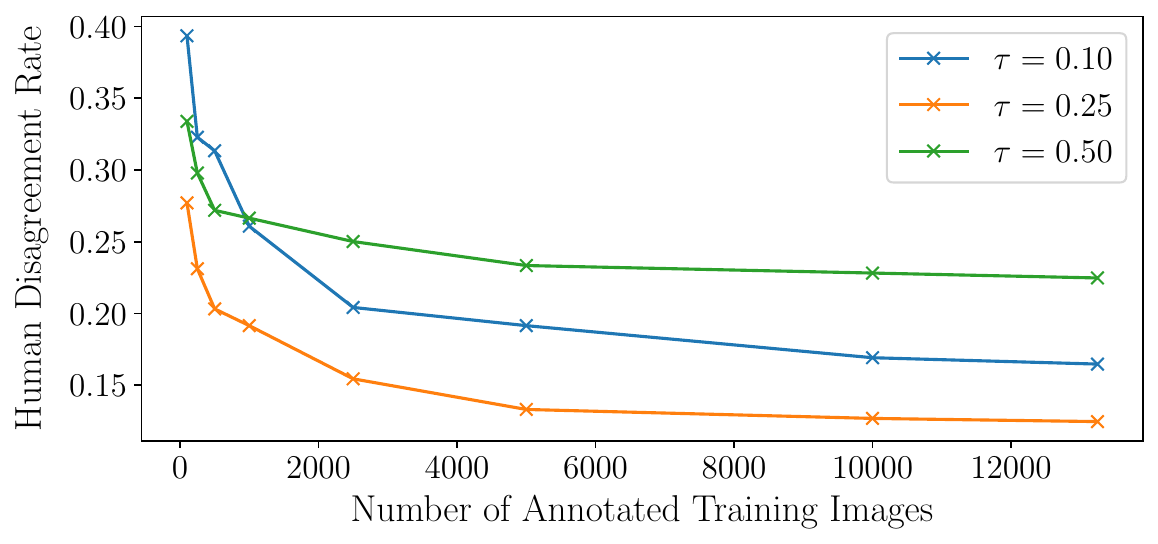}
  \vspace{-5pt}
  \caption{Validation set $\text{HDR}_\tau$ at various thresholds as a function of the number of annotated training images.}
  \label{fig_scalingstudy}
  \vspace{-15pt}
\end{figure}

\begin{figure*}[tp]
  \vspace{1pt}
  \centering
  \includegraphics[width=7in]{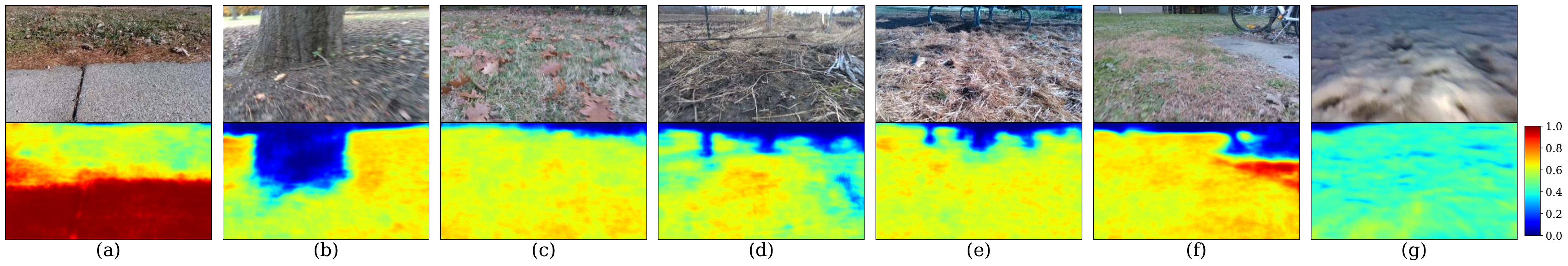}
  \vspace{-5pt}
  \caption{Example inference visualizations from our validation split of the WayFAST dataset \cite{wayfast}. The input color images are shown in the top row, and the corresponding traversability predictions are shown in the bottom row. The traversability score ranges from 0 to 1 for each image, with a higher traversability score indicating that a region is more easily traversed.}
  \label{fig_qualitative1}
  \vspace{-15pt}
\end{figure*}

As labeling efficiency is a critical motivation behind our work, in Fig. \ref{fig_scalingstudy} we analyze the effect of the number of annotated training images on the validation set $\text{HDR}$. For this analysis, we use the \textit{MT+PT} model with $\mathcal{L}_{\text{RIZZ}}$ loss (where each training image has 3 annotated pixels). The results in Fig. \ref{fig_scalingstudy} demonstrate that increasing the number of annotated images greatly improves $\text{HDR}$ when the number of annotated images is small. As the size of the training dataset grows, the effect of adding more annotated images decreases. For example, training on $5000$ or $10000$ images shows similar $\text{HDR}$ to training on the entire $13248$ training images, indicating that labeling the entire dataset is not crucial for high performance. However, this behavior will likely vary for different datasets (e.g., datasets with low complexity or low image variation may require fewer annotations for high performance compared to more complex datasets).

\subsection{Qualitative Results}\label{sec:qual_results}
We visualize several validation set images predicted using W-RIZZ (using the \textit{MT+PT} variant with $\mathcal{L}_{\text{RIZZ}}$ loss) in Fig.~\ref{fig_qualitative1}. The results shown in Fig. \ref{fig_qualitative1} demonstrate that W-RIZZ provides accurate relative traversability predictions. Our method also captures nuanced notions of traversability due to its relative labeling, like the smooth concrete being more traversable than bumpy grass in Fig.~\ref{fig_qualitative1}(a) and Fig. \ref{fig_qualitative1}(f). 
Despite using only three annotated pixels per training image, W-RIZZ can also capture small obstacles like the metal post in Fig. \ref{fig_qualitative1}(d) and the legs of the table in Fig. \ref{fig_qualitative1}(e).

\begin{figure}[bp!]
  \vspace{-10pt}
  \centering
  \includegraphics[width=3.2in]{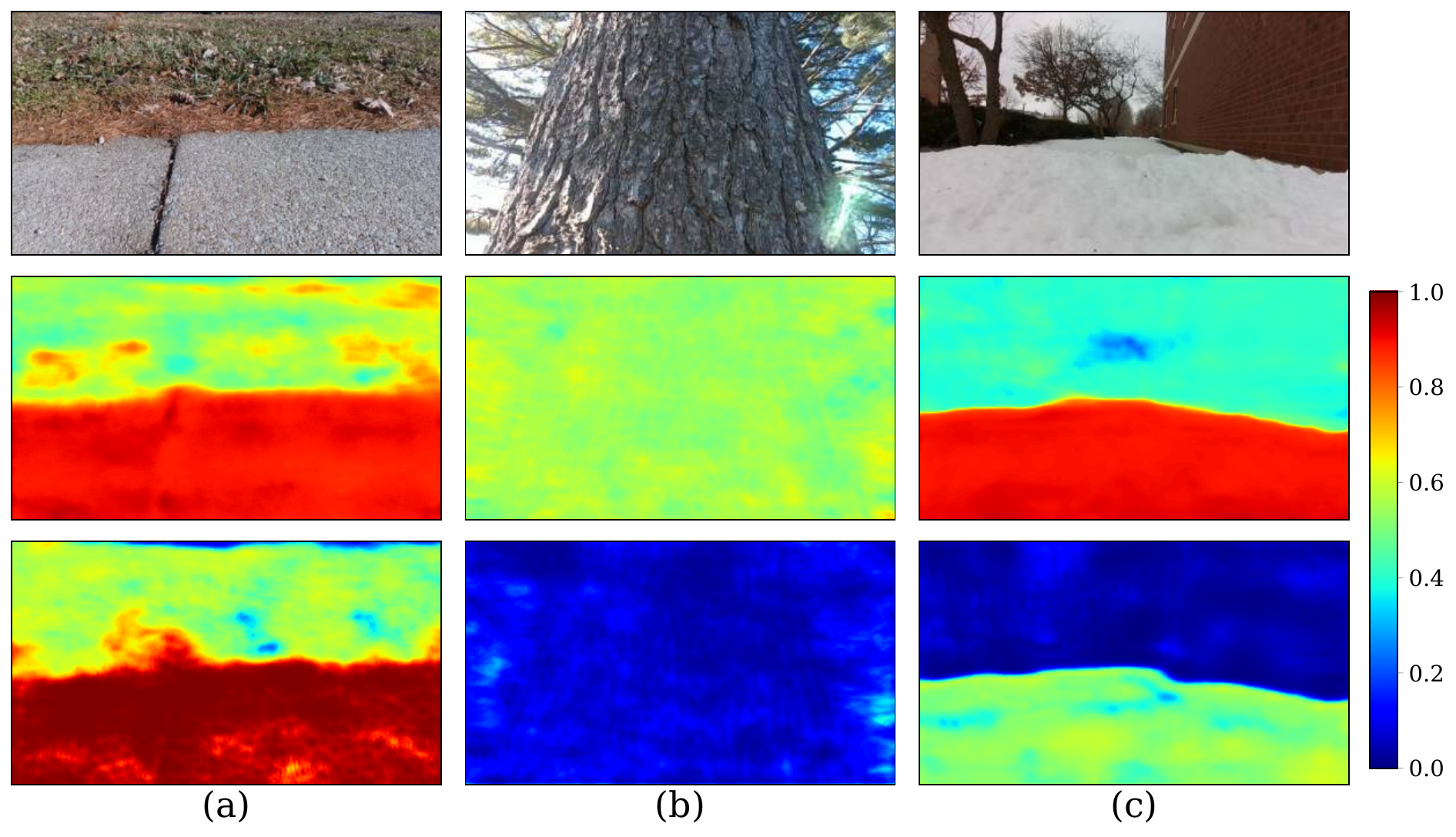}
  \vspace{-5pt}
  \caption{Predictions on images from the WayFAST dataset \cite{wayfast}, showing color images (top), as well as predictions from a model trained without cross-image labeling (middle) and with cross-image labeling (bottom).}
  \label{fig_qualitative3}
\end{figure}

In Fig. \ref{fig_qualitative3}, we show predictions when using only intra-image labeling and when using cross-image and intra-image labeling. For these predictions, we followed the same training procedure as used for the results in Table III. The results in Fig. \ref{fig_qualitative3} demonstrate that using cross-image labeling significantly improves prediction consistency. For example, if only intra-image labeling is used, the snow mound in Fig.~\ref{fig_qualitative3}(c) has approximately the same traversability score as the smooth concrete in Fig.~\ref{fig_qualitative3}(a), and the untraversable tree in Fig.~\ref{fig_qualitative3}(b) has approximately the same traversability score as the traversable grass in Fig.~\ref{fig_qualitative3}(a). However, when cross-image labeling is incorporated, the resulting traversability predictions are significantly more consistent.

\begin{figure}[bp!]
  \vspace{-10pt}
  \centering
  \includegraphics[width=3.4in]{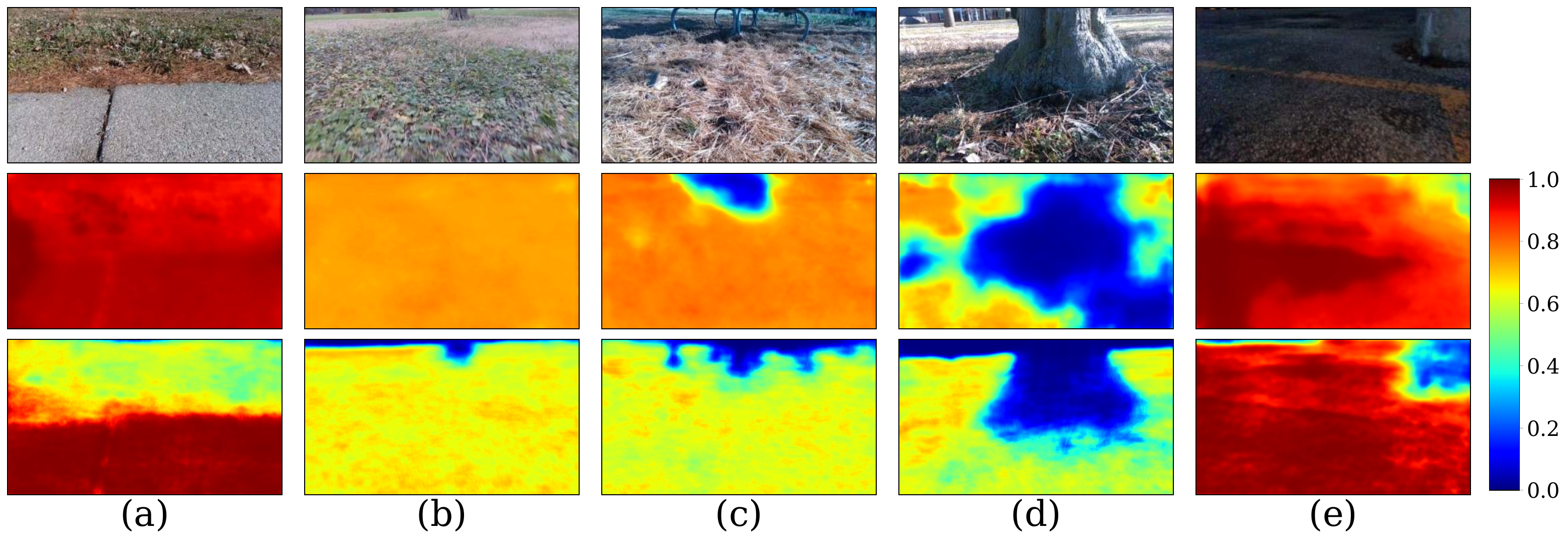}
  \vspace{-5pt}
  \caption{Example predictions on images from the WayFAST dataset \cite{wayfast}, showing color images (top), predictions from WayFAST (middle), and predictions from W-RIZZ (bottom).}
  \label{fig_qualitative_wayfast}
\end{figure}

In Fig. \ref{fig_qualitative_wayfast} we provide predictions on sample images from the WayFAST dataset \cite{wayfast} when trained using WayFAST and W-RIZZ, where we train WayFAST as described in Sec. \ref{sec:quant_results}.
While our method uses a strategy of weakly-supervised manual labeling (with only 3 pixels per image labeled), WayFAST labels traction coefficients in a self-supervised manner and produces many more annotated pixels per image. Although W-RIZZ requires human annotation, our annotation strategy is efficient and takes only seconds per image. Our method also does not require any data beyond images, whereas WayFAST assumes a kinodynamic model and requires additional data/models to predict traction coefficients. Finally, WayFAST requires a robot to experience traction conditions to label them, which makes it difficult to safely label untraversable zones.

The results provided in Fig. \ref{fig_qualitative_wayfast} highlight important distinctions between the predictions from WayFAST and W-RIZZ. For example, W-RIZZ predicts higher traversability for concrete than grass (which is beneficial as smooth concrete is likely a more appealing surface to traverse than bumpy grass if all else is equal), while WayFAST fails to make a distinction between the two surfaces. In addition, the difficulty of labeling untraversable areas with the self-supervised strategy leads to erroneous traversability predictions in Fig. \ref{fig_qualitative_wayfast}(b)-(e). For example, WayFAST fails to capture the distant obstacle in Fig. \ref{fig_qualitative_wayfast}(b), the right leg of the metal table in Fig. \ref{fig_qualitative_wayfast}(c), and the concrete pillar in Fig. \ref{fig_qualitative_wayfast}(e). WayFAST also produces odd, overly-cautious predictions regarding the tree in Fig. \ref{fig_qualitative_wayfast}(d). By comparison, W-RIZZ produces significantly more accurate predictions for obstacles.

\subsection{Comparison to Strongly-Supervised Segmentation}
We additionally compare our sparse annotation method to strongly-supervised segmentation on the RUGD dataset~\cite{rugd}. This dataset consists of images which are labeled for semantic segmentation using 25 classes (if the void class is included) relevant to outdoor navigation in unstructured environments, and we divide its semantic annotations into 4 traversability tiers: tier 3 (concrete and asphalt), tier 2 (dirt, sand, grass, gravel, mulch, and rock-bed), tier 1 (water and bush), and tier 0 (all other classes), where low tiers correspond to lower traversability. We use these tiers to create automatic annotations for W-RIZZ (with 3 annotated pixels per image) which we use to train our model. We use automated annotations to ensure consistent comparison to the strongly-supervised labeling. As over $90\%$ of the pixels in the top half of training set images correspond to sky or tree, we bias our sampling such that $90\%$ of samples must be in the bottom half of the image (with the remaining $10\%$ being unrestricted) to provide more informative labels. We train the \textit{MT} W-RIZZ model (i.e., we do not pretrain on RUGD) using the sparse automatically-generated annotations, and compare our method to a strongly-supervised model using the same network architecture (where the last layer is modified to produce 25-class segmentation outputs).

We provide results for our method and the strongly-supervised approach on the RUGD test set in Table IV. We compute our metrics assuming 4-class segmentation of the traversability tiers described earlier, and report mean intersection-over-union ($\textit{mIOU}$) and mean pixel-wise accuracy ($\textit{mAcc}$), where the means are computed over the 4 classes. We also provide frequency-weighted metrics where the mean is weighted by the relative frequency of each class ($\textit{fw-mIOU}$ and $\textit{fw-mAcc}$). To convert the traversability scores from W-RIZZ to discrete traversability tiers, we compute the cutoff that differentiates tier $N$ from tier $N-1$ ($N \in \{3,2,1\}$) as $\text{cutoff}_N = \frac{(\mu_N - \sigma_N) + (\mu_{N-1} + \sigma_{N-1})}{2}$, where $\mu_N$ and $\sigma_N$ are the mean and standard deviation of the W-RIZZ traversability scores for tier $N$ on the training set. The segmentation model produces 25-class outputs, and we remap these to the 4 traversability tiers for our metric calculations.

The results in Table IV highlight that our sparse annotation strategy shows reasonably high performance despite using very few labeled pixels in training. Even though only three annotated pixels are provided for each training image, W-RIZZ shows results within $4\%$ of the strongly-supervised segmentation approach for each metric, despite the strongly-supervised model being trained with significantly more annotations (as the strongly-supervised segmentation model is trained using images where every pixel is labeled).

\begin{table}[tp]
\vspace{5pt}
\caption{W-RIZZ vs. Strongly-Supervised Segmentation on RUGD}
\vspace{-8pt}
\begin{center}
\begin{tabular}{@{} l *{4}{r} @{}}
\toprule
{Method} & \multicolumn{1}{c}{$\text{mIOU}$} & \multicolumn{1}{c}{$\text{fw-mIOU}$} & \multicolumn{1}{c}{$\text{mAcc}$} & \multicolumn{1}{c}{$\text{fw-mAcc}$} \\
\midrule

{W-RIZZ} & \multicolumn{1}{c}{0.668} & \multicolumn{1}{c}{0.764} & \multicolumn{1}{c}{0.767} & \multicolumn{1}{c}{0.856} \\

{Segmentation} & \multicolumn{1}{c}{0.692} & \multicolumn{1}{c}{0.783} & \multicolumn{1}{c}{0.785} & \multicolumn{1}{c}{0.879}\\
\bottomrule
\end{tabular}
\end{center}
\vspace{-15pt}
\end{table}

\subsection{Real-World Experiments}
\label{sec:result_onrobot}
We further analyze our method by conducting real-world experiments on a TerraSentia robot (a skid-steer robot developed by EarthSense Inc.). For these experiments, we utilize the traversability predictions from W-RIZZ with a non-linear model predictive controller (NMPC) described by Gasparino \textit{et al.}~\cite{wayfast}.
The neural network inference and controller are run on the robot's onboard computer (Jetson AGX Orin) at $15 \ \text{Hz}.$ While we use the controller introduced by Gasparino \textit{et al.}~\cite{wayfast} for these experiments, W-RIZZ is agnostic to the control strategy and can be used with other controllers that consume traversability prediction images.

We compare our approach against the following baselines:
\begin{enumerate}
    \item \textit{LiDAR}: an obstacle-avoiding approach based on purely geometric information acquired by a 2D LiDAR unit.
    \item \textit{WayFAST} \cite{wayfast}: an RGB-only variant of a state-of-the-art self-supervised visual traversability prediction method.
    \item \textit{WayFAST*} \cite{wayfast}: an enhanced version of WayFAST trained on a different dataset and using the RGB-D WayFAST network architecture.
\end{enumerate}

The W-RIZZ and WayFAST models are trained with the same neural network architecture (which uses only RGB images) on the full dataset from Gasparino \textit{et al.} \cite{wayfast}. We train W-RIZZ using our relative traversability labels, and we train WayFAST using the labels from the WayFAST dataset.

The robot used for our experiments had a camera setup that differed slightly from that of the original WayFAST dataset (the WayFAST dataset has a different camera that is angled more towards the ground). While this did not cause issues with W-RIZZ, it led to poor performance of the original WayFAST model. Thus, we provide results with WayFAST*, which is trained on a new dataset that more closely matches the camera setup of the robot used for the experiments and that uses the RGB-D network architecture from WayFAST (which utilizes RGB and depth images). The W-RIZZ model used in these experiments is only trained on the original WayFAST dataset and does not use a depth camera.

\begin{figure}[tp]
  \vspace{2pt}
  \centering
  \includegraphics[width=2.6in]{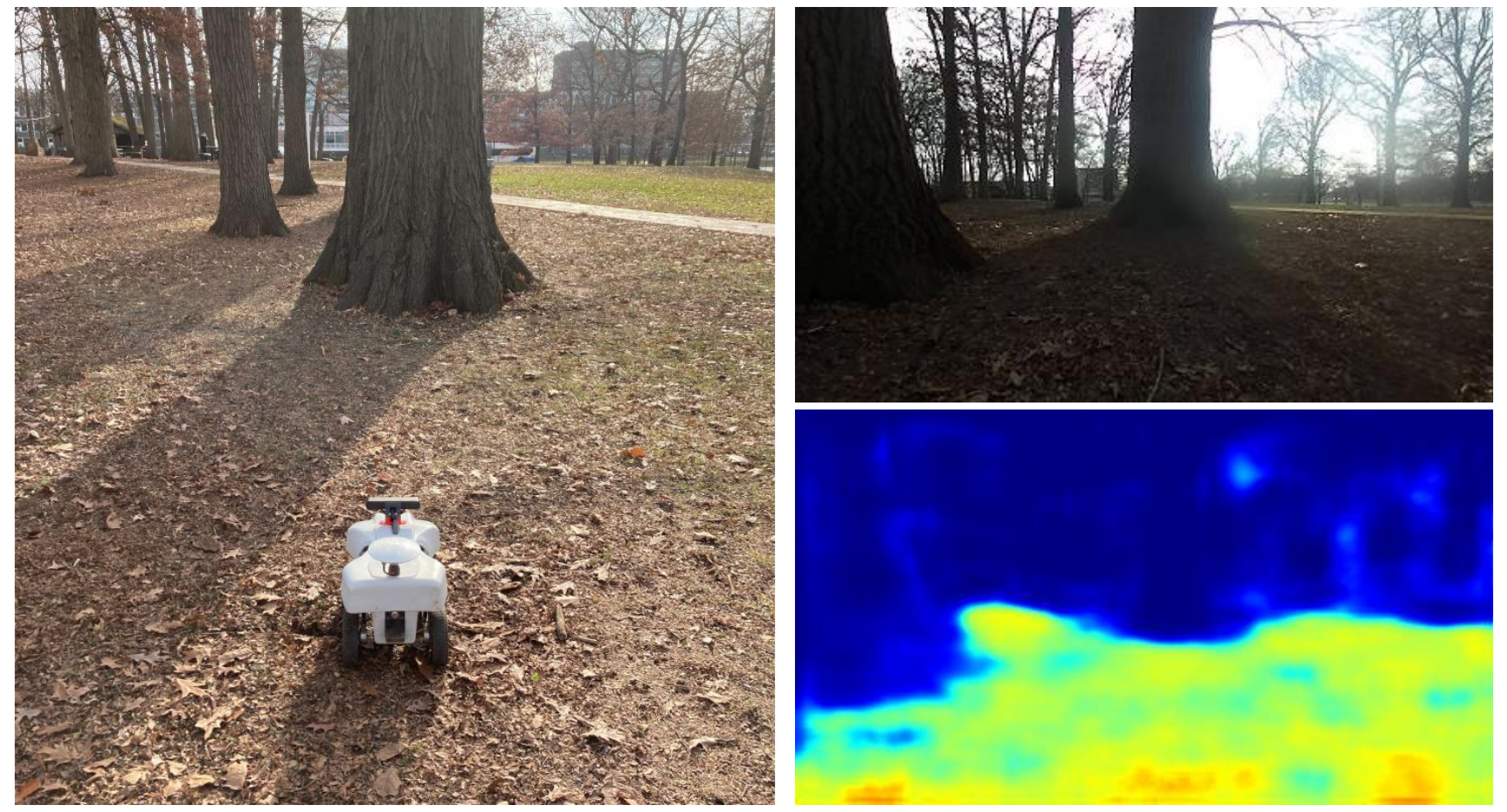}
  \caption{
  Navigation experiment setup, showing the TerraSentia robot (left), as well as a sample camera image from the robot (top right) and the associated traversability prediction from W-RIZZ (bottom right).}
  \label{fig_field_experiments}
\end{figure}

We perform our navigation experiments in a grove of trees where a robot must navigate approximately $20 \ \text{m}$ between obstacles to reach a goal; the setup of the experiment is shown in Fig. \ref{fig_field_experiments}. We perform $5$ navigation trials for each method, and the results are shown in Table V. All methods took approximately the same time to reach the goal (if the goal was reached). Videos from the experiments and for a longer navigation task are provided as supplemental material.

The results from Table V demonstrate that both the LiDAR baseline and W-RIZZ are able to successfully navigate the course in all attempts. The WayFAST model (trained on the same data as W-RIZZ) crashed into a tree in every navigation attempt. These results suggest greater generalization capability with W-RIZZ when compared to WayFAST.
The results for WayFAST* show improved performance compared to WayFAST, demonstrating that incorporating depth and retraining on a new dataset help to overcome the domain shift. However, even when given such training advantages over W-RIZZ, WayFAST* still showed lower navigation success rate than W-RIZZ (in one of its navigation attempts WayFAST* failed to avoid an obstacle, crashing into a tree).

\begin{table}[tp]
\caption{Robot Navigation Experimental Results}
\vspace{-8pt}
\begin{center}
\begin{tabular}{@{} l *{3}{r} @{}}
\toprule
Method & \multicolumn{1}{c}{Successful Trials} & \multicolumn{1}{c}{Total Trials} & \multicolumn{1}{c}{Success Rate}\\
\midrule
LiDAR & \multicolumn{1}{c}{5} & \multicolumn{1}{c}{5} & \multicolumn{1}{c}{100\%} \\
WayFAST & \multicolumn{1}{c}{0} & \multicolumn{1}{c}{5} & \multicolumn{1}{c}{0\%} \\
WayFAST* & \multicolumn{1}{c}{4} & \multicolumn{1}{c}{5} & \multicolumn{1}{c}{80\%} \\
W-RIZZ & \multicolumn{1}{c}{5} & \multicolumn{1}{c}{5} & \multicolumn{1}{c}{100\%} \\
\bottomrule
\end{tabular}
\end{center}
\vspace{-15pt}
\end{table}

While both W-RIZZ and LiDAR show $100\%$ success rate in our experiments, we note the difference in the perception method of RGB camera versus 2D LiDAR. The 2D LiDAR has a greater field of view and produces simpler measurement data that only provide a purely geometric view of the environment, which can fail for traversable obstacles like tall grass and cannot capture features like smoothness of terrain. The richer vision-based sensing modality used by W-RIZZ can capture such nuances as shown in Sec. \ref{sec:qual_results}. The 2D LiDAR can also display measurement limitations, as obstacles above or below the measurement plane of the LiDAR unit cannot be detected.

\section{Conclusion}

We have introduced a novel method of relative traversability estimation that uses a small number of sparsely annotated point pairs which are labeled with annotations of relative traversability. We enhance the performance of our method by introducing a new loss for relative traversability estimation and a novel cross-image labeling scheme. We have validated our approach through offline evaluation in various settings (including snow, tall grass, forests, and semi-urban areas) and through deployment on an actual mobile robot platform.

Although our method does not depend on the robotic platform that is used, it shows the limitation of requiring relabeling of data if a robot's capabilities change significantly. In addition, while our method still requires manual effort for labeling, the sparse relative annotation scheme is simple, does not need extensive training, and demands significantly less effort than labeling for strongly-supervised segmentation (while maintaining high accuracy). Our use of manual relative traversability labels also circumvents the need to define semantic classes and does not assume constant traversability characteristics within semantic classes (in contrast to semantic segmentation approaches), allowing for direct prediction of a continuous traversability score while not requiring robot experience or the data, models, or assumptions needed in self-supervised approaches. Our framework can be used with any neural network for dense prediction tasks and is compatible with techniques like pretraining to further improve results.

Our online experiments demonstrate that our method can outperform a state-of-the-art self-supervised approach and shows greater robustness to domain shifts like varying camera angles.
Possible future work includes incorporating additional methods for weakly-supervised learning (e.g., pseudo-labeling) and combining our method with foundation models to improve accuracy and generalization performance.

\ifCLASSOPTIONcaptionsoff
  \newpage
\fi

\bibliographystyle{IEEEtran}
\bibliography{references}

\end{document}